\pdfoutput=1

\documentclass[letterpaper, 10 pt, conference]{ieeeconf}  

\IEEEoverridecommandlockouts                              

\usepackage{tikz}
\usepackage{mathtools}
\usepackage{amsfonts}
\usepackage{amsmath}
\usepackage{amssymb}
\usepackage{bbm}
\usepackage{caption}
\usepackage{subcaption}
\usepackage{algorithm}
\usepackage{algpseudocode}
\usepackage{tabularx}
\usepackage[skip=1ex]{caption}

\usepackage{fixltx2e}

\DeclareCaptionType{copyrightbox} 
\newcommand{\note}[1]{}

\usepackage{tikz,tkz-base}
\usetikzlibrary{shapes,decorations,shadows}
\usetikzlibrary{decorations.pathmorphing}
\usetikzlibrary{decorations.shapes}
\usetikzlibrary{fadings}
\usetikzlibrary{patterns}
\usetikzlibrary{calc}
\usetikzlibrary{decorations.text}
\usetikzlibrary{decorations.footprints}
\usetikzlibrary{decorations.fractals}
\usetikzlibrary{shapes.gates.logic.IEC}
\usetikzlibrary{shapes.gates.logic.US}
\usetikzlibrary{fit,chains}
\usetikzlibrary{positioning}
\usepgflibrary{shapes}
\usetikzlibrary{scopes}
\usetikzlibrary{arrows}
\usetikzlibrary{backgrounds}

\tikzset{latent/.style={circle,fill=white,draw=black,inner sep=1pt,
  minimum size=20pt, font=\fontsize{10}{10}\selectfont},
  obs/.style={latent,fill=gray!25},
  const/.style={rectangle, inner sep=0pt},
  factor/.style={rectangle, fill=black,minimum size=5pt, inner sep=0pt},
>={triangle 45}}

\pgfdeclarelayer{b}
\pgfdeclarelayer{f}
\pgfsetlayers{b,main,f}

\newcommand{\bD}{\textbf{D}}

\newcommand{\blambda}{\boldsymbol\lambda}

\newcommand{\bzeta}{\boldsymbol\zeta}

\newcommand{\DONE}[1]{}

\title{\LARGE \bf Bayesian Nonparametric Modeling of Driver Behavior using HDP Split-Merge Sampling Algorithm}

\author{Vadim Smolyakov$^{1}$ and Julian Straub$^{2}$ and Sue Zheng$^{3}$ and John W. Fisher III$^{4}$
\thanks{$^{1}$Vadim Smolyakov is with the Department of Electrical Engineering and Computer Science, Massachusetts Institute of Technology, Cambridge, MA 02139, USA
        {\tt\small vss@csail.mit.edu}}%
\thanks{$^{2}$Julian Straub is with the Department of Electrical Engineering and Computer Science, Massachusetts Institute of Technology, Cambridge, MA 02139, USA
        {\tt\small jstraub@csail.mit.edu}}%
\thanks{$^{3}$Sue Zheng is with the Department of Electrical Engineering and Computer Science, Massachusetts Institute of Technology, Cambridge, MA 02139, USA
        {\tt\small zhengs@mit.edu}}%
\thanks{$^{4}$John W. Fisher III is with the Faculty of the Computer Science and Artificial Intelligence Laboratory, Massachusetts Institute of Technology,
        Cambridge, MA 02139, USA
        {\tt\small fisher@csail.mit.edu}}%
}

\newcommand\blfootnote[1]{%
  \begingroup
  \renewcommand\thefootnote{}\footnote{#1}%
  \addtocounter{footnote}{-1}%
  \endgroup
}

\begin{document}
\maketitle
\thispagestyle{empty}
\pagestyle{empty}
\blfootnote{This work was partially funded by the Ford-MIT alliance and by the Office of Naval Research Multidisciplinary Research Initiative (MURI) program, award N00014-11-1-0688.}
\begin{abstract}
Modern vehicles are equipped with increasingly complex sensors. These sensors generate large volumes of data that provide opportunities for modeling and analysis.  Here, we are interested in exploiting this data to learn aspects of behaviors and the road network associated with individual drivers.  Our dataset is collected on a standard vehicle used to commute to work and for personal trips.  A Hidden Markov Model (HMM) trained on the GPS position and orientation data is utilized to compress the large amount of position information into a small amount of road segment states. Each state has a set of observations, i.e. car signals, associated with it that are quantized and modeled as draws from a Hierarchical Dirichlet Process (HDP). The inference for the topic distributions is carried out using HDP split-merge sampling algorithm. The topic distributions over joint quantized car signals characterize the driving situation in the respective road state. In a novel manner, we demonstrate how the sparsity of the personal road network of a driver in conjunction with a hierarchical topic model allows data driven predictions about destinations as well as likely road conditions.

\end{abstract}
\begin{figure}
  \centering
  \includegraphics[trim=50 150 125 5, clip, width=0.23\textwidth]{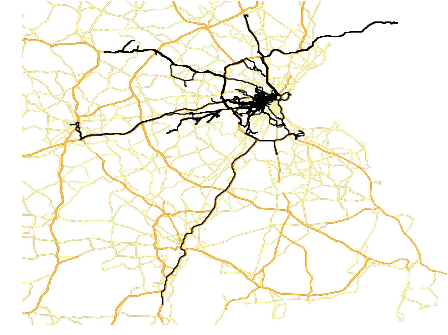}
  \includegraphics[trim=5 20 5 45, clip, width=0.23\textwidth]{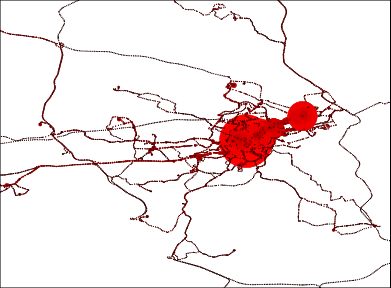}
  \hspace{0.3cm}

  \includegraphics[trim=5 50 5 45, clip, width=0.40\textwidth]{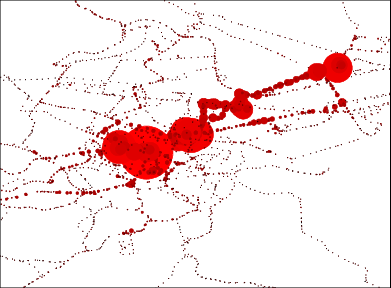}

\caption{Spatial distribution of the observations in the car dataset -- zooming in from left to right. The leftmost plot shows the sparsity of the road states the driver is visiting compared to the full road network. The middle and right plot depict the distribution of the number of measurements taken in the respective road segment -- the redder and larger the more. Clearly, this distribution is very imbalanced. A small number of states account for a majority of the measurements. \label{fig:roadstateStats}} 
\end{figure}
\section{Introduction}
%
Vehicles are equipped with an increasing number of sensors and electronics to react dynamically to changing road conditions and to increase driver safety. As a result, large volumes of driver-specific data related to driving conditions and driver behavior are generated.  We are interested in analyzing this data to learn models of driving behavior. Such models could be used to anticipate dangerous situations, to improve the driving schedule of a person, and to tailor various aspects of the driving experience to the individual.

\begin{figure*}
\centering
\begin{subfigure}[c]{0.47\textwidth}
  \centering
  \begin{tikzpicture}
    \draw [red,  thick] (0,1.5) circle[radius = .5];
    \draw [red,  thick] (2,1.5) circle[radius = .5];
    \draw [red,  thick] (4,1.5) circle[radius = .5];

    \node [red] at (0,1.5) {$x_0$};
    \node [red] at (2,1.5) {$x_1$};
    \node [red] at (4,1.5) {$x_2$};

	\draw [ thick, ->] (0.5,1.5) -- (1.5,1.5);
	\draw [ thick, ->] (2.5,1.5) -- (3.5,1.5);
	\draw [ thick, ->] (4.5,1.5) -- (5.5,1.5);	
 
    \node at (6,1.5) {$\cdots$};

    \draw [blue,  thick] (0,0) circle[radius = .5];
    \draw [blue,  thick] (2,0) circle[radius = .5];
    \draw [blue,  thick] (4,0) circle[radius = .5];
    
    \node [blue] at (0,0) {$y_0$};
    \node [blue] at (2,0) {$y_1$};
    \node [blue] at (4,0) {$y_2$};

	\draw [thick, ->] (0,1) -- (0,.5);
	\draw [thick, ->] (2,1) -- (2,.5);
	\draw [thick, ->] (4,1) -- (4,.5);	
    
  \end{tikzpicture}
\end{subfigure}
\begin{subfigure}[c]{0.47\textwidth}
  \centering
  \begin{tikzpicture}[scale=0.17]
\draw [gray, ultra thick] plot [smooth] coordinates { (-3,0) (0,0) (3,3) (9,3) (18,6) (18,9)};


\draw [red, ultra thick, rotate=0 ] (-4,0) circle[x radius = 3.5, y radius = 5];
\node [red] at (-3,6) {$a$};

\draw [red, ultra thick,shift={(1.5,1.5)}, rotate=45] circle[x radius = 2.1, y radius = 0.9];
\node [red] at (1.5,4) {$b$};

\draw [red, ultra thick, rotate=0 ] (6.15,3) circle[x radius = 3, y radius = 1.5];
\node [red] at (6.15,5.5) {$c$};

\draw [red, ultra thick,shift={(13.5,4.5)}, rotate=20] circle[x radius = 4.5, y radius = 1.5];
\node [red] at (13.5,7.2) {$d$};

\draw [red, ultra thick,shift={(19,9)}, rotate=20] circle[x radius = 2.5, y radius = 4.5];
\node [red] at (15.5,12) {$e$};

\node [red] at (5,10) {hidden states};

\node [blue] at (12,-3) {GPS measurements};
\draw [blue, thick, fill = blue ] (-1.5,-.4) circle(.3cm);

\draw [blue, thick, fill = blue ] (0.8,1.1) circle(.3cm);

\draw [blue, thick, fill = blue ] (1.8,1.8) circle(.3cm);
\draw [blue, thick, fill = blue ] (5.0,3.4) circle(.3cm);
\draw [blue, thick, fill = blue ] (6.0,3.5) circle(.3cm);
\draw [blue, thick, fill = blue ] (7.5,2.8) circle(.3cm);
\draw [blue, thick, fill = blue ] (10.5,3.8) circle(.3cm);
\draw [blue, thick, fill = blue ]  (12,3.7) circle(.3cm);
\draw [blue, thick, fill = blue ]  (14,4.5) circle(.3cm);
\draw [blue, thick, fill = blue ] (16,6) circle(.3cm);
\draw [blue, thick, fill = blue ] (18,8) circle(.3cm);
  \end{tikzpicture}
\end{subfigure}
\caption{(Left) A standard HMM where the hidden variables (in red) correspond to road segments and the observation variables (in blue) include position and heading measurements. (Right) A conceptual rendering of the HMM. The physical road is shown in gray while the HMM representation of the road states is shown in red. Position measurements are shown in blue. 
\label{fig:roadstates}}
\end{figure*}
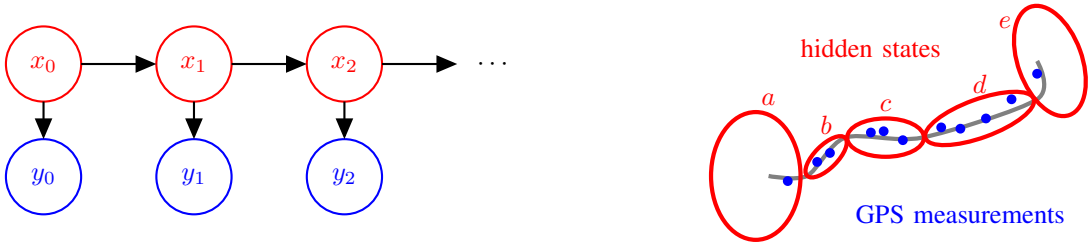
Here, we use data collected from one vehicle's sensors over numerous trips to construct a Hierarchical Dirichlet Process (HDP) model of driving behavior and road conditions. HDPs are commonly used for topic modeling of text corpora \cite{teh2006hierarchical,chang14hdp,wang2011online,hoffman2013stochastic} to uncover the set of topics that comprise each document in the corpus. In our case, the documents are road segments and the words are associated quantized sensor measurements. The topics in the HDP model are sensor distributions in the road segments; these distributions capture the driving conditions in each road segment as encountered by the driver as well as their driving behavior and common driving conditions. To our knowledge this is a new approach for modeling driving behavior. Unlike related work which is based on assumptions about the capabilities and behaviors of humans (i.e. see for an overview \cite{Ranney1994733}), our model is purely data driven.

It is important to note that the hierarchy within the HDP model allows sharing of measurements across similar road segments. This is an appealing aspect of the model since it enables us to learn an expressive model for road segments which are visited rarely via similar road segments that are visited more often. 
In order to utilize an HDP model, we first organize the sensor data into "documents" (i.e., road segments and their associated quantized measurements). We consider the case in which a road map is not available, however, it is straightforward to incorporate such information. Additionally, typical drivers often traverse a small subset of the roads in the road network. We use a Hidden Markov Model (HMM) to learn the road segments. The HMM condenses position information from recorded trips into road segment states. The set of hidden states effectively corresponds to a sparse road network which consists only of the roads which the driver has traversed. We then use the trained HMM to associate sensor measurements to road segments to produce "documents" for the HDP model.

In addition to organizing the data for the HDP model, the HMM also provides insight into driver behavior such as typical routes and probable destinations.  Special hidden states are introduced in the HMM to represent starting locations (sources) and destinations. Conseqently, identification of the most likely route between two states and finding the distribution over probable destinations become well-posed questions and allow us to make route and destination predictions.

The contributions of this paper are (1) to show how sparsity in the HMM transition matrix together with starting and absorption states lead to accurate long term predictions of driver routes and destinations and (2) the novel application of a HDP split-merge sampler to model the joint distribution of quantized vehicle signals scalable to a large number of road-segments.
%
%
%


%
\section{Hidden Markov Model}
An HMM is used to model the trips that a driver takes through a road network. We explore two models for the HMM. In the first model, the hidden state corresponds to a road segment, a start location, or a destination. In this model, the future path is independent from the past path when conditioned on the current road segment. We expect this to be a poor model of driver behavior since this is likely an oversimplification; the past can provide considerable information about the future. For instance, drivers often do not return to a previously visited state within a trip (unless they are lost). 

In our second model, we attempt to capture more of the trip history in the current state by augmenting the road states with the start location. Under this model, the road segment at the next time instance depends only on the current road segment and the start location. We will show that this model is more representative of driver behavior and provides accurate predictions of destinations and routes. We describe this second model below. The first model is a simplification of the described model. 

\begin{figure*}[htb]
\centering
\includegraphics[width=0.45\textwidth]{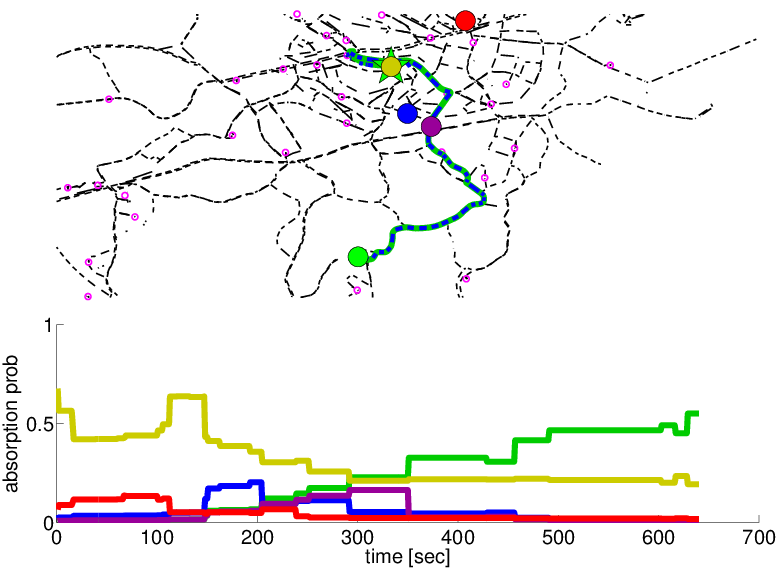}
\includegraphics[width=0.45\textwidth]{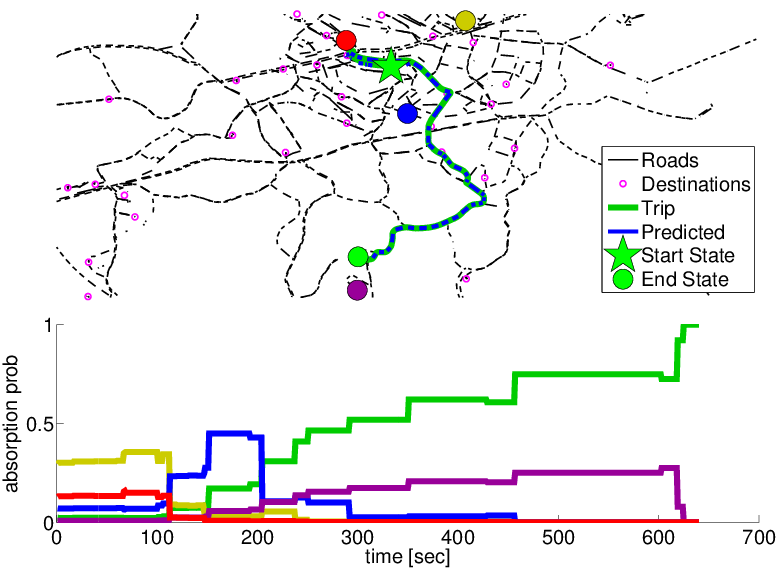}
\caption{Predicted routes and absorption probabilities corresponding to the five most likely destinations of the model without (left) and with (right) start locations. The most likely destinations differ between the two models; notably, in the left model, there is significant probability that the driver will return to his starting location (shown in yellow) at the start of the trip, while the starting location is not a likely destination in the model augmented with start location. Furthermore, the absorption probability of the true destination, shown in green, dominates over alternative possible destinations sooner in the bottom model.}
\label{fig:CompareHMMs}
\end{figure*}
\subsection{Hidden States}
Each hidden variable, $x_t$, in the HMM (see Fig. \ref{fig:CompareHMMs}) takes on a value from the set of hidden states, $\mathcal{X}$.  Source states, $\mathcal{X}_S$, destination states, $\mathcal{X}_D$, and road segment states augmented by the source state, $\mathcal{X}_R \times \mathcal{X}_S$, compose the set of hidden states: $\mathcal{X} = \mathcal{X}_S \cup \mathcal{X}_D \cup (\mathcal{X}_R \times \mathcal{X}_S)$. Destination states are absorbing states which are indicated by key-off events in the data. Similarly, source states are indicated by key-on events. The distribution over the initial state, $x_0 \in \mathcal{X}$, is parameterized as $p(x_0 = m) = \theta_m$. Conditioned on the current state, $x_t$, the distribution for the next state, $x_{t+1}$, is parameterized as 
\[ p(x_{t+1} = m|x_t=k) = \theta_{km}.\]
Since physically realizable transitions occur only between road segments in close proximity, we would expect most transition probabilities to be zero. We use a Dirichlet prior on the parameters with $\alpha < 1$ to favor a sparse transition matrix: 
\[p(\theta_{k1}, \theta_{k2}, \ldots, \theta_{k \vert \mathcal{X} \vert}) \propto \prod_{i=1}^{\vert \mathcal{X} \vert} \theta_{ki}^{\alpha-1}.\]

\subsection{Observation Model}
Each trip contains measurements of position, $r_t$ and heading, $h_t$. When the vehicle has GPS, the recorded position is the GPS position; otherwise the reported position is obtained by dead reckoning, a process which estimates position by combining the previous position with aggregated incremental changes in a relative coordinate system. Positions which are inferred using dead reckoning are indicated by an inferred-position indicator, $q_t$; for such measurements, we model a larger uncertainty associated with the measurement. 

In addition to position and heading measurements, there is a key-on event at the start of each trip which indicates that hidden state must be from the set of source states. In the measurement model, we have a binary key-on indicator variable, $k^{on}_t$, which takes value $1$ if a key-on event occurs at time $t$. Similarly, there is a key-off event and corresponding indicator variable, $k^{off}_t$, which indicates that the hidden state is from the set of destination states. This set of measurements comprise the observation $y_t = \{r_t, h_t, q_t, k^{off}_t, k^{on}_t\}$. 
Conditioned on the hidden state, the measurement model is as follows:
\begin{align*}
p(r_t, h_t, q_t, k^{on}_t, k^{off}_t| x_t) = &p(r_t|q_t, x_t)p(h_t|x_t) p(q_t|x_t)\\
& \times p(k^{on}_t|x_t)p(k^{off}_t|x_t). 
\end{align*}
Note that the conditional distribution for position depends on the value on the inferred-position indicator; a larger uncertainty is associated with the position when the position has been inferred. Position is Gaussian with state-dependent parameters: 
\begin{equation*}
p(r_t|q_t, x_t) = \begin{cases} \mathcal{N}(r_t; \mu_{r,x_t}, \Sigma_{r,x_t}) &q_t = 0 \\
\mathcal{N}(r_t; \mu_{r,x_t}, c \cdot \Sigma_{r,x_t}) &q_t = 1 \end{cases}
\end{equation*}
where $c > 1$ is a constant used to capture the increase in uncertainty of the inferred position. 

Heading is also Gaussian with its own state-dependent parameters:
\[p(h_t|x_t) = \mathcal{N}(h_t; \mu_{h,x_t}, \Sigma_{h,x_t}).\]
The inferred position indicator has a Bernoulli distribution with parameter, $p_{x_t}$. The key-on and key-off measurements are indicators of source and destination states respectively: $k^{on}_t = \mathbbm{1}\left(x_t \in \mathcal{X}_S\right)$, $k^{off}_t = \mathbbm{1}\left(x_t \in \mathcal{X}_D\right)$ and have degenerate distributions.

We would expect measurements arising from the same physical location (road segment) to have parameters which do not depend on the source state. Therefore, the measurement parameters are independent of the source state when conditioned on the road segment state. That is, $\mu_{r,x_t} = \mu_{r, x_r}$ for $x_t \in \{x_r \times \mathcal{X}_S\}$, where $x_r \in \mathcal{X}_R$, and likewise for the other measurement parameters. When we estimate the measurement parameters for a road segment, this formulation allows us to aggregate observations from trips which start at different locations but share this physical road. 

Similarly, there will be pairs of source and destination states will correspond to the same physical location. If a trip ends at a given location, the next trip will typically start from the same location. Since this pair of states share physical properties, these states will share measurement parameters. 

\subsection{EM Updates}
Given the volume of data under consideration, we find that an EM formulation using explicit state assignments provides a tractable learning approach. This approach yields locally optimal values for the set of parameters $\psi = \{\theta_m, \theta_{km}, \mu_{r,m}, \Sigma_{r,m}, \mu_{h,m}, \Sigma_{h,m}, p_m\}$ for $m,k \in \mathcal{X}$ using measurements from $N$ trips. The EM updates consist of iteratively finding the most likely assignment for the hidden states given previous parameter estimates, then using these assignments to improve the parameter estimates. The reader is referred to  \cite{dempster1977em} for an introduction to the EM algorithm. 

To initialize the parameters, we run DP means \cite{jordan2012dpmeans} to cluster the measurements based on position and heading; a state is created from each cluster. DP means allows us to initialize the model without pre-specifying the number of states. Measurements assigned to the cluster (state) are used to calculate initial values for measurement model parameters. The transition matrix is initialized as a full matrix with higher probability for states which are closer together. The distribution for the first state is initialized as a uniform distribution. 

\subsection{Predicting Routes and Destinations}
Using the HMM model, we can predict a driver \textit{route} from state $a$ to state $b$ by identifying the sequence of states $\{x_1^*=a, x_2^*, \ldots, x_N^*=b\}$ with the highest likelihood:
\[\prod_{i=1}^{N-1} p(x^*_{i+1} | x^*_i)  = \max_n \max_{\underset {x_1 = a, \: x_n = b}{\{x_1, \ldots, x_n\} \in \mathcal{X}}} \prod_{i=1}^{n-1}  p(x_{i+1} | x_i) \]
It is well known, \cite{simmons2006driverRoutes}, that this can be formulated as a shortest path problem by defining a graph on the hidden states with edge weights $w_{ij} = -\log p(x_j | x_i)$. 

Additionally, from any road segment state, $i$, we can find the probability of reaching any destination state, $j$; this is known as the absorption probability. The absorption probability, $a_{ij}$, is the probability of reaching absorbing state, $j$, if the chain starts from state, $i$, and can be found by solving the following set of equations: 
\begin{align*}
a_{jj} &= 1 \quad \forall j \in \mathcal{X}_D \\
a_{ji} &= 0 \quad \forall j \in \mathcal{X}_D  \quad \forall i \neq j \\
a_{ij} &= \theta_{ij} + \sum_{k \in \mathcal{X} \setminus j} \theta_{ik}a_{kj} \quad \forall j \in \mathcal{X}_D, \: \forall i \in \mathcal{X}_R
\end{align*}
This gives us a probability distribution over destinations when we start from a given road state.

\newcommand{\ho}{\text{ho}}
\newcommand{\obs}{\text{obs}}
\newcommand{\who}{w_\ho}
\newcommand{\phiho}{\phi_\ho}
\newcommand{\wobs}{w_\obs}

\subsection{Bayesian Nonparametric Topic Modeling of Car Signals}
%

Thus far, we have formulated an HMM model for driving behavior which has predictive aspects. The model is also used to organize the data into "documents" so that we can perform HDP topic modeling on the dataset. 
In this section we discuss how we combine a standard HDP model with the use of an HMM to discover documents. We also relate our HDP model for car signals to the classical HDP topic model.

To bridge the gap between the classical HDP topic modeling of text corpora and the modeling of car signals, such as velocity, acceleration and rotational speed, note the following correspondences: 
\begin{align*}
\text{word} &\leftrightarrow \text{car signals at one instance in time}\\ \text{document} &\leftrightarrow \text{road segment}\\ 
\text{corpus} &\leftrightarrow \text{map}
\end{align*}

Each learned road segment from the HMM is used as a document in the HDP model. To obtain a set of sensor measurements associated with a road segment, or a set of words from the document, we perform ML assignment of road states for the trips and assign the corresponding sensor measurements to those road states.

Since the car signals are continuous quantities, we quantize them using DP means \cite{jordan2012dpmeans} and use a discrete base measure equivalent to the classical text corpus topic model.
This means that words are described by a multidimensional vector, which amounts to modeling the joint distribution over all signals.

In the next section we briefly summarize the HDP model and describe how we use a parallelizable HDP split-merge sampler to perform inference on a large number of discretized car measurements. For a more detailed presentation the reader is referred to the original papers \cite{teh2006hierarchical} and \cite{chang14hdp}. 

\subsection{HDP model}
The HDP model describes a set of $D$ documents, which contain $N_j$ words $x_{ji}$ each. In this context, the documents correspond to trips and the words in a document correspond to quantized sensor measurements from the trip. The distribution of words $x_{ji}$ is modeled as a mixture of topic distributions $\theta_k$. 

\begin{figure}\label{fig:gm_hdp}
  \centering
  \includegraphics[width=0.6\textwidth, trim={10 150 10 150}]{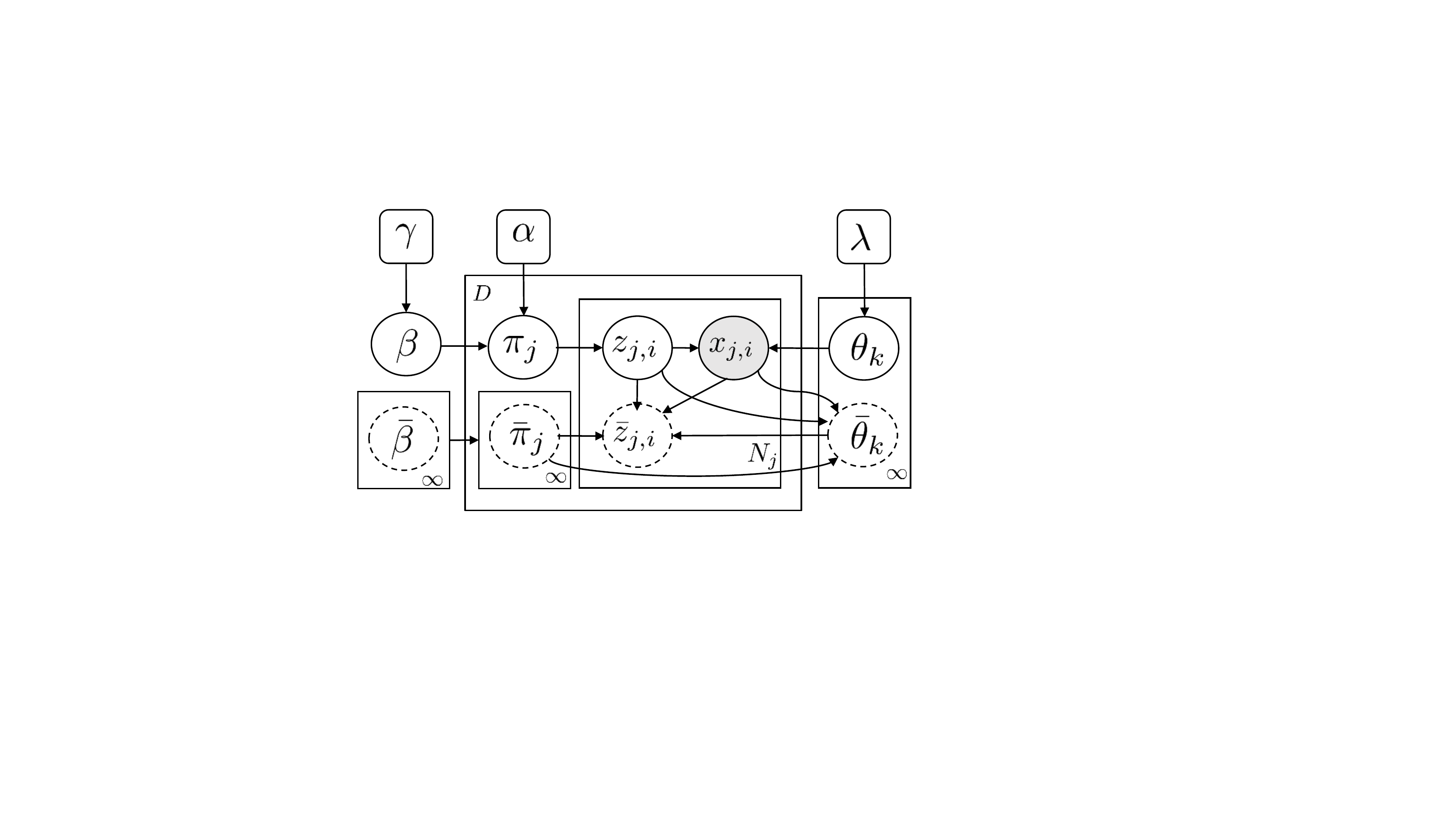}
  \caption{Direct Assignment graphical model representation of an HDP \cite{teh2006hierarchical} augmented with split-merge nodes \cite{chang14hdp}.}
\end{figure}

\begin{table*}[t]
\centering
\begin{tabularx}{\linewidth}{XX}
    \includegraphics[width=\columnwidth]{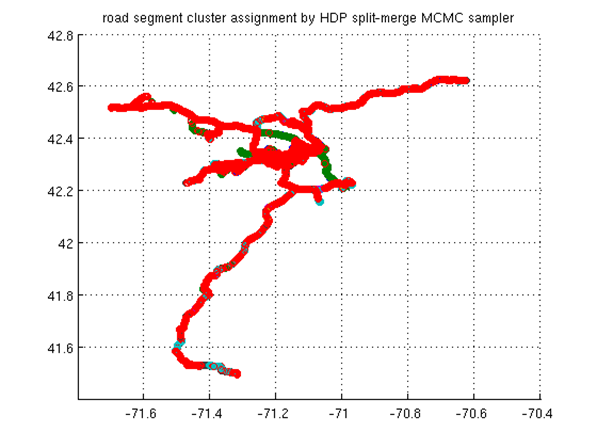}
    \label{fig:hdp_results1}
&
    \includegraphics[width=\columnwidth]{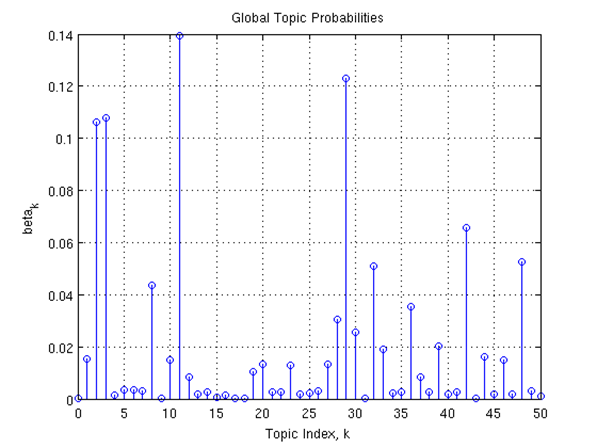}
    \label{fig:hdp_results2}
\\
    \includegraphics[width=\columnwidth, height=6.4cm]{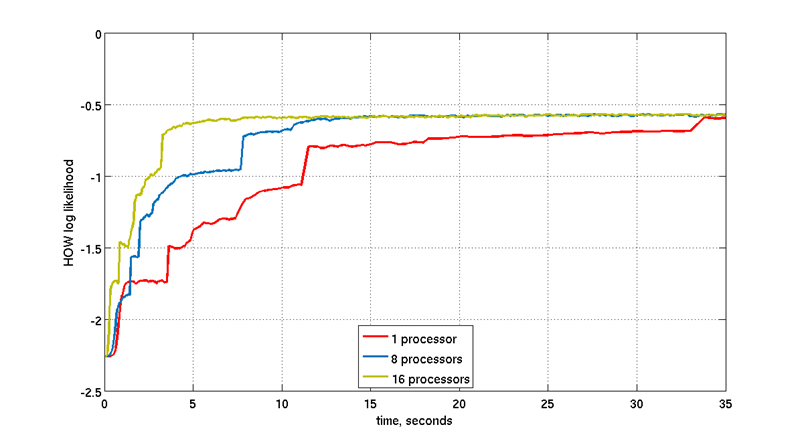}
    \label{fig:hdp_results3}
&
    \includegraphics[width=\columnwidth]{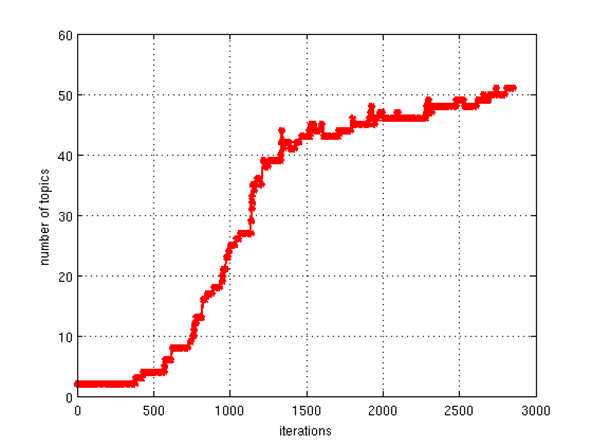}
    \label{fig:hdp_results4}
\\
\end{tabularx}
\end{table*}

\begin{figure*}
  \centering 
\begin{subfigure}[c]{\textwidth}
  \centering 
  \includegraphics[width=\textwidth]{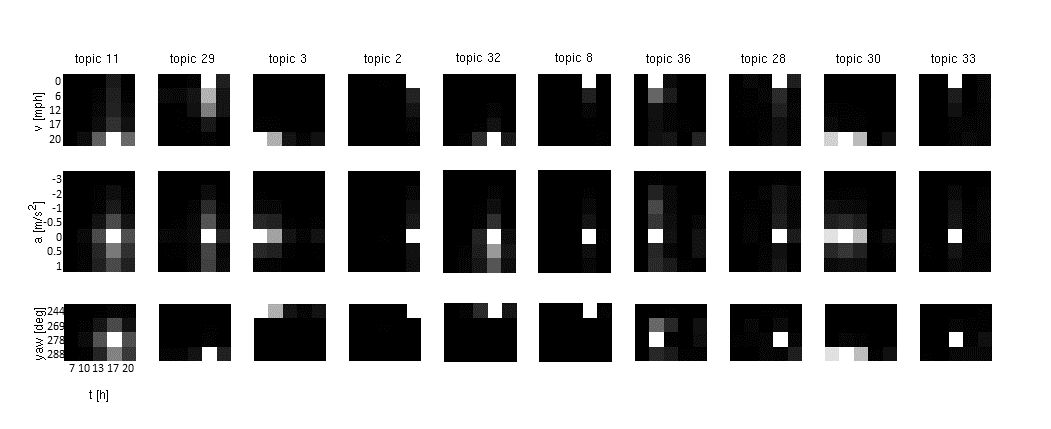}
\end{subfigure}
  \caption{Plots in the first row show the topic assignments (left) and global topic distribution (right). The plots in the second row show held-out word (HOW) log-likelihood (left) and number of topics vs iterations (right). Finally, the top 10 likely topics inferred by HDP split-merge sampler are shown at the bottom row for different car signals.}
\end{figure*}


The graphical model can be described as follows. At the top-level, we have $\beta \sim \mathrm{GEM}(1,\gamma)$ that determines the global topic proportions via the stick-breaking construction \cite{teh2006hierarchical}. Also, shared at the top-level are the global word distributions $\theta_k \sim f_{\theta}(\theta;\lambda)$. At the document-level, $\pi_j$ is the DP with $\beta$ as the base measure: $\pi_j \sim \mathrm{DP}(\alpha, \beta)$, from which topic assignment labels $z_{ji} \sim \mathrm{Cat}(\pi_j)$ are sampled. Finally, the observed signals can be expressed as $x_{ji} \sim f_x(x_{ji};\theta_{z_{ji}})$, where the topic assignment $z_{ji}$ indexes corresponding word distribution $\theta_{z_{ji}}$. The dashed nodes are the auxiliary split-merge nodes that learn a two-component mixture model for each cluster. These "sub-clusters" are then used to propose splits and merges that are selected over time. This method combines a Gibbs sampler that is restricted to non-empty clusters with a Metropolis-Hastings (MH) algorithm that proposes sub-cluster splits and merges.\\

\subsection{Split-Merge HDP Sampler}
Because of the large quantity of data, we use highly parallelizable split-merge HDP sampler described in \cite{chang14hdp}. Augmenting the sample space with sub-clusters leads to proposals of likely splits and merges. A combination of a restricted Gibbs sampler (that does not create new clusers) with split/merge moves results in an ergodic Markov chain. 

\subsubsection{Resricted Gibbs Sampler}
Let $m_{jk}$ denote the number of clusters in document $j$ with shared topic $k$ and let $n_{jtk}$ denote the number of words in document $j$ in cluster $t$ with topic $k$. Then, the marginal counts $n_{j..}=\sum_{t,k}n_{jtk}$ and $m_{j.}=\sum_{k}m_{jk}$ represent the number of words and topics in document $j$, respectively. Extending the DA sampling algorithm results in the following restricted posterior distributions:
\begin{eqnarray}\label{equ:eqn15}
  p(\beta|m)&=&\mathrm{Dir}(m_{.1},...,m_{.K},\gamma)\\
  p(\pi_j|\beta,z)&=&\mathrm{Dir}(...,\alpha \beta_K+n_{j.K},\alpha \beta_{K+1})\\
  p(\theta_k|x,z) &\propto& f_x(x_{I_{k}};\theta_k)f_{\theta}(\theta_k;\lambda)\\
  p(z_{ji}|x,\pi_j,\theta) &\propto& \sum_{k=1}^{K}\pi_{jk}f_x(x_{ji};\theta_k)1[z_{ji}=k]\\
  p(m_{jk}|\beta,z) &=& f_m(m_{jk};\alpha \beta_k, n_{j.k})\\
  &=&\frac{\Gamma (\alpha \beta_k)}{\Gamma(\alpha \beta_k + n_{j.k})}s(n_{j.k},m_{jk})(\alpha \beta_k)^{m_{jk}}\nonumber
\end{eqnarray}
Since $p(\beta|m)$ is not known analytically, we use the auxiliary variable $m_{jk}$. $s(n,m)$ denotes unsigned Stirling numbers of the first kind. Note that the last components $\beta_{K+1}$ and $\pi_{j(K+1)}$ aggregate the weight of all empty topics. Finally, $I_k=\{j,i; z_{ji}=k\}$ denotes the set of indices in topic $k$, and $f_x$ and $f_{\theta}$ denote the observation and prior distributions. The equations above can be sampled in parallel and fully specify the restricted Gibbs sampler. 

The method combines a Gibbs sampler that is restricted to non-empty clusters with a Metropolis-Hastings (MH) algorithm that proposes splits and merges.\\

\subsubsection{Subcluster Splits and Merges}
For each topic $k$, we fit two sub-topics $kl$ and $kr$ referred to as the left and right sub-clusters. Each topic is augmented with global sub-topic proportions $\bar{\beta}_k = \{\bar{\beta}_{kl},\bar{\beta}_{kr}\}$, document-level sub-topic proportions $\bar{\pi}_{jk}=\{\bar{\pi}_{jkl},\bar{\pi}_{jkr}\}$, and sub-topic parameters $\bar{\theta}_k = \{\bar{\theta}_{kl},\bar{\theta}_{kr}\}$. Moreover, each word $x_{ji}$ is associated with sub-topic assignment $\bar{z}_{ji}\in \{l,r\}$. Then the marginal posterior distributions can be derived \cite{chang14hdp} as:
\begin{eqnarray}\label{equ:eqn610}
  p(\bar{\beta}_k|\cdot)&=&\mathrm{Dir}(\gamma + \bar{m}_{.kl},\gamma + \bar{m}_{.kr})\\
  p(\bar{\pi}_{jk}|\cdot)&=&\mathrm{Dir}(\alpha \bar{\beta}_{kl}+\bar{n}_{j.kl},\alpha \bar{\beta}_{kr}+\bar{n}_{j.kr})\\
  p(\bar{\theta}_{kh}|\cdot) &\propto& f_x(x_{I_{kh}};\bar{\theta}_{kh})f_{\theta}(\bar{\theta}_{kh};\lambda)\\
  p(\bar{z}_{ji}|\cdot) &\propto& \bar{\pi}_{jk}f_x(x_{ji};\bar{\theta}_k)1[\bar{z}_{ji}=k]\\
  p(\bar{m}_{jkh}|\cdot) &=& f_m(\bar{m}_{jkh};\alpha \bar{\beta}_{kh}, \bar{n}_{j.kh})
\end{eqnarray}
Notice the similarity between these equations and ones derived earlier. Inference is performed by interleaving the sampling equations $(1)-(5)$ with marginal posterior equations $(6)-(10)$.

\subsubsection{Metropolis-Hastings}

A Metropolis-Hastings framework proposes splits and merges of sub-clusters and either accepts or rejects them. Let $v=\{\beta,\pi,z,\theta\}$ and $\bar{v}=\{\bar{\beta},\bar{\pi},\bar{z},\bar{\theta}\}$ be a set of regular and auxiliary variables, respectively. Then a sampled proposal $\{\hat{v},\hat{\bar{v}}\}\sim q(\hat{v},\hat{\bar{v}}|v)$ is accepted with probability:
\begin{equation}\label{equ:mh}
   P_a = \min \big[1, \frac{p(x,\hat{v})p(\hat{\bar{v}}|x,\hat{v})}{p(x,v)p(\bar{v}|x,v)}\cdot \frac{q(v|x,\hat{v})q(\bar{v}|x,\hat{\bar{v}},v)}{q(\hat{v}|x,v)q(\hat{\bar{v}}|x,\bar{v},\hat{v})} \big]
\end{equation}

\begin{algorithm}
\caption{HDP Split-Merge Algorithm}
\label{alg:minibatch}
\begin{algorithmic}[1]
\State Propose assignments $\hat{z}$, global proportions $\hat{\beta}$, document proportions $\hat{\pi}$ and parameters $\hat{\theta}$
\State Defer proposal of auxiliary variables to restricted sampling of $(1)$-$(10)$ 
\State Accept / reject the proposal with the Hastings ratio. 
\end{algorithmic}
\end{algorithm}

\begin{figure*}[h!]
  \centering
  \includegraphics[trim=0 11 0 110, clip, width=0.49\textwidth]{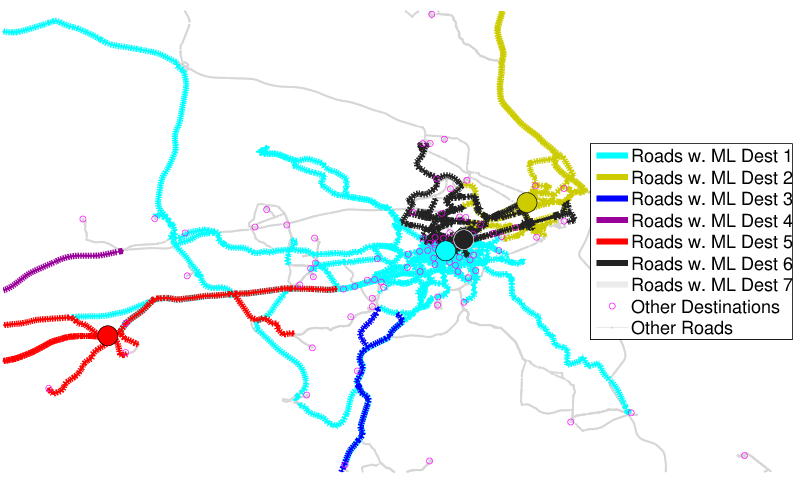}
  \includegraphics[trim=0 11 0 110, clip, width=0.49\textwidth]{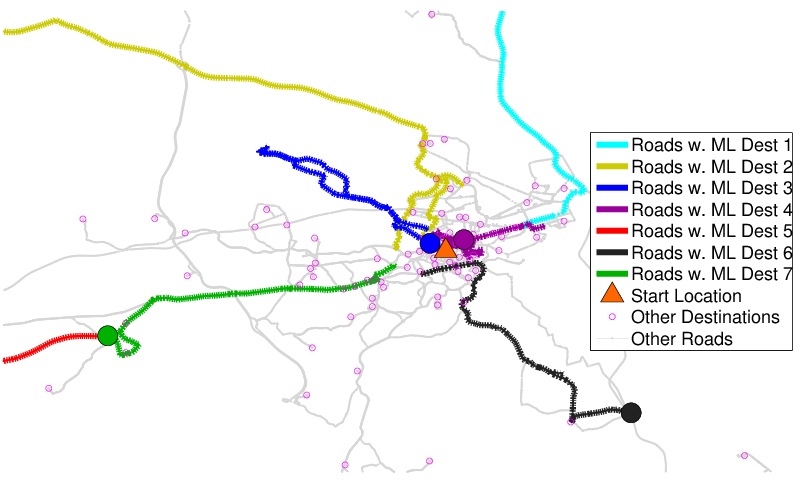}
  \caption{Most likely destination for each road segment without (left) and with (right) start location. The seven most popular destination locations are indicated by large colored circles. In the model augmented with start location, the start location for this plot is shown in red. Only roads whose most likely destination belongs to the set of seven most popular destinations in each model are plotted in the color corresponding to the destination. In the right plot, we see that the model is able to capture the phenomenon that the most likely destination is not a destination which the driver has already passed, or equivalently, the destination will not be on any typical path between the start location and the road segment}
  \label{fig:AbsorbForEachRoad}
\end{figure*}
\subsection{Performance Evaluation}
%
%
%
To evaluate the performance of the HDP model, we are computing the average log predictive probability of held-out words. To compute this probability, we split a test document into two sets: held out words $w^\ho$ and observed words $w^\obs$. Then we update the model using the observed words. This gives us the posterior parameters $\{\bzeta^\obs,\blambda^\obs \}$ for the test document which we in turn use to find $\phi^\ho$ for the held-out words. Now we can compute the probability of a held-out words as given all training data $\bD$ as well as the observed words $w^\obs$ in this document: 
\begin{equation}
  \begin{array}{l}
    p(w^\ho_n| \bD, w^\obs) = \sum^T_{i=1} q(z_{n} = i|\phi^\ho)  \\
    \sum^K_{k=1} q(c_{i} = k| \bzeta^\obs) p(w^\ho_n | z_{n} = i,c_{i} =k,\blambda^\obs) 
\end{array}
\end{equation}
where $p(w^\ho_n| z_{n}^i,c_{i}^k,\blambda^\obs) = \frac{\blambda^\obs_{k}(w^\ho_n)}{\sum_w \blambda_k^\obs(w) }$ is the conditional distribution of a held-out word under the posterior distribution of words in this document. 

As a model to compare the HDP to, we utilize a non-hierarchical model that assumes a Categorical distribution with a Dirichlet prior for the words in each road-state. These distributions are modeled completely independent -- not connected via a hierarchy like in the HDP model. This allows us to compute posterior Categorical distributions given the observed words in each road-state.


%
\section{Results}
In the following we will first give results for the predictive power of the HMM model before we describe a topic model for the joint distribution of speed and time-of-day measurements.

\subsection{Dataset Description}
Our dataset comprises of 1K trips recorded from a standard car used by a single
driver. The routes are mostly commuting to work but also some longer range
trips outside the city.
 
The GPS position and heading measurements of the car are used to train the HMM
model. From various other signals of the car we selected quantized car velocity
and time of day for the HDP topic model. These were selected, since they
contain interesting information both about the driving behavior as well as the
driving situation in a road state.

\subsection{Predicting Routes and Destinations}
To evaluate the quality of the learned HMM, we examine the ability of the HMM to predict the destination for 20 held-out trips under the two different models. Additionally, we compare the path of the held-out trips against the most likely route obtained from the transition matrix for the HMM.


Fig.~\ref{fig:CompareHMMs} shows the performance of the two models on a held-out trip. The plots under the maps in the figure show the absorption probabilities for the probable destinations as a function of time. The maps above show the trip, the most likely route between the source and destination state, and the locations of the probable destinations. While the most likely path between source and destination from both models agrees with the observed trip trajectory, we observe that the augmented model is able to identify the correct destination sooner than the first model. In fact, the first model is able to correctly predict the destination after 10\% of the trip for only 3 of the held-out trips while the augmented model is able to do so for 11 of the trips. 

In Fig.~\ref{fig:AbsorbForEachRoad}, we show the most likely destination for each road segment. For the augmented model, since each state associated with a road segment also has a start location, we've chosen a particular start location to illustrate the differences between these two models. In particular, when starting from the specified start location, we see that trips which traverse beyond destination 7 in Fig.~\ref{fig:AbsorbForEachRoad} (bottom) are more likely to terminate at a destination which is further from the starting location. The unaugmented model is unable to make this distinction, so trips which traverse road segments near destination 5 in Fig.~\ref{fig:AbsorbForEachRoad} on the top (which corresponds to destination 7 in Fig.~\ref{fig:AbsorbForEachRoad} on the bottom) are likely to terminate at that destination.


The results show that the most likely route obtained by the models frequently align exactly with the path of the held-out trips. This can be explained through the sparsity of the transition matrices; since each state can only transition to few states, and very often just one state, long term predictions in this model are quite accurate.

\subsection{HDP Model}
\begin{figure*}
  \centering 
  \begin{subfigure}[c]{0.47\textwidth}
    \centering 
    \includegraphics[trim=5 130 5 25, clip, width=\textwidth]{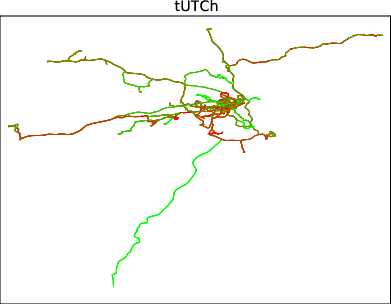}
    \caption{ML estimate of time-of-day per road state computed from empirical distribution. Color-coded from green (early in the day) to red (late at night).\label{fig:empMarginal_tod}}
  \end{subfigure}
  \hspace{0.3cm}
  \begin{subfigure}[c]{0.47\textwidth}
    \includegraphics[trim=5 130 5 25, clip, width=\textwidth]{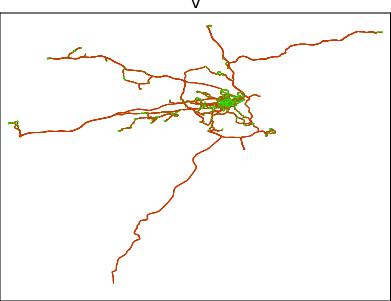}
    \caption{ML estimate of speed per road state computed from empirical distribution. Color-coded from green (slow speeds) to red (fast driving). \label{fig:empMarginal_v}}
    \centering 
  \end{subfigure}

  \begin{subfigure}[c]{0.47\textwidth}
    \centering 
    \includegraphics[trim=5 130 5 25, clip, width=\textwidth]{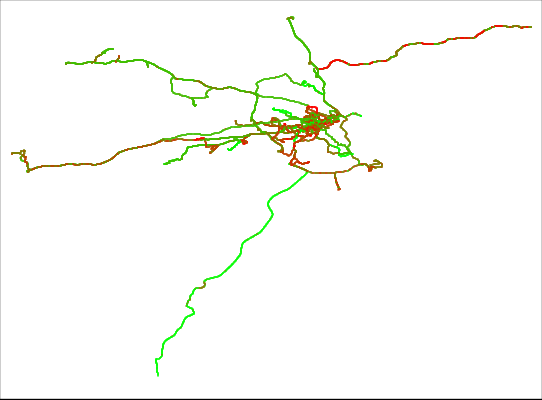}
    \caption{ML estimate of time-of-day per road state computed via inferred HDP model.\label{fig:hdpMarginal_tod}}
  \end{subfigure}
  \hspace{0.3cm}
  \begin{subfigure}[c]{0.47\textwidth}
    \includegraphics[trim=5 130 5 25, clip, width=\textwidth]{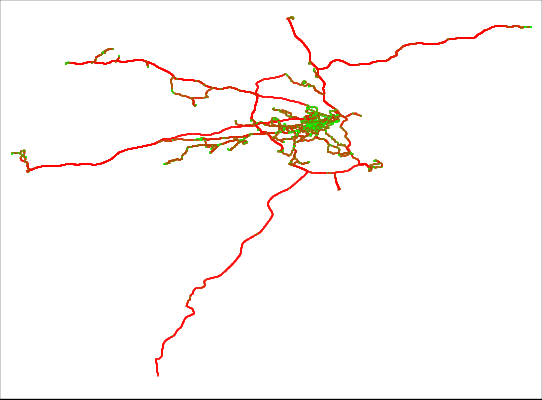}
    \caption{ML estimate of speed per road state computed via inferred HDP model. \label{fig:hdpMarginal_v}}
  \end{subfigure}
  \caption{Plots of the Maximum Likelihood (ML) estimates of time-of-day and speed computed from the inferred HDP model (bottom row) and the empirical distribution (top row).\label{fig:marginals}}
\end{figure*}
We are quantizing velocity and time-of-day measurements to words that can be fed into the HDP inference algorithm.
Quantization is performed via DP k-means clustering over the individual signals. 

The speed measurements arrive at a rate of 1~Hz from the GPS sensor. 
There are 696k joint observations -- velocity/time-of-day pairs -- across all
12k road states. As can be seen in Fig.~\ref{fig:roadstateStats}, these
observations are distributed non-uniformly -- we get a lot of measurements on
the daily commute route and few on highways leading outside the city. This
means that the road-state corpus has very imbalanced document sizes when
compared to text corpus modeling. However, our results demonstrate, that this
presents no issue to the inference algorithm.

We empirically found the following set of parameters: $\gamma=10.0$ and $\alpha=0.1$, corresponding to the global and local concentration parameters, respectively. 

Fig. 5 demonstrates that the hierarchy in the HDP is
able to pool measurements from different road-states to obtain a descriptive
topic for these. For each road state we obtain the maximum likelihood (ML)
topic assignment and plot the respective road states in red. This pooling of
observations can for example be observed for topics~0~and~41, which consist of
almost all highway road states as can be seen in the ML topic assignment plots
(compare the red road segments to the highways depicted in map in
Fig.~\ref{fig:roadstateStats}).

Using the inferred mixture of topics for each state, we can now compute the ML
estimate of the marginals for the individual sensor signals and plot them
color-coded for each road state. Fig.~\ref{fig:marginals} shows this for the
marginal
over speed and time-of-day.  Comparing the spatial distribution of the ML speed
estimates computed from the inferred HDP model in Fig.~\ref{fig:hdpMarginal_v} with ML estimates obtained from the empirical
distribution depicted in Fig.~\ref{fig:empMarginal_v}, we can see that
the HDP model is able to capture the distribution of the input data.  

Additionally, it is clear from the spatial distribution of the ML estimates of driving speeds, that the HDP
model captures for example the fact, that inner city driving is slower than
highway driving. The ML estimates of time-of-day (Fig.~\ref{fig:hdpMarginal_tod} and \ref{fig:empMarginal_tod}) show that the trips outside the city
were not undertaken in the morning or evening.

\note{
\subsection{Comparison of Inferred HDP Marginals with Empirical Marginals}
The combined HMM and HDP models can be used to obtain expected road conditions along various routes. In this example, we look at the speed profile along a typical driver route. The HMM is used to find the most likely route between two locations. The HDP provides the marginal distribution over speed for each road state. 

Since the HDP pools measurements from the different road-states, we expect the inferred HDP marginals to be a more robust estimate than an empirical marginal distribution. Especially for road-states have only a few measurements associated with them, an empirical marginal distribution of the sensor measurements based on just the data associated to that road-state can be a poor estimate of the marginal. We compare the two marginals using the data likelihood of a held-out set under the marginals. 

Since the empirical distribution over sensor measurements has a zero likelihood for measurement values which haven't been previously observed at the road-state, we apply a Dirichlet distribution prior and use the predictive distribution as the empirical marginal. The likelihood of  speed measurements conditioned on time-of-day for the 20 held-out trips is shown in Fig.~\ref{fig:CompareMarginals}. 
}

%
\section{Conclusion}
We have shown that the inherent sparsity of the learned personal road network allows accurate long term predictions of driver routes. Additionally, augmenting the model with start location yields a more representative model which provides better destination predictions. 
Exploiting the hierarchy of the HDP topic model, we are able to learn expressive topic distributions despite the fact that the number of car signal measurements differs widely between different road states. 
The combination of both types of of models allows us to model the driving behavior of an individual driver. This type of model can for example assist in optimizing the daily commute route or help predict traffic jams. 
As a next step it would be interesting to compare the driver models for different drivers to allow driver classification based on the driving behavior.
%

%
%
%
%
\bibliographystyle{./IEEEtran} 
\bibliography{jstraub}
\end{document}